\documentclass[hidelinks]{article}
\usepackage{arxiv}
\usepackage{nicefrac}       
\usepackage{microtype}      
\usepackage{booktabs}       
\usepackage{doi}
\usepackage{adjustbox}
\usepackage{caption}
\usepackage{float}
\usepackage[T1]{fontenc}
\usepackage{graphicx}
\usepackage{hhline}
\usepackage[utf8]{inputenc}
\usepackage{multicol}
\usepackage{multirow}
\usepackage{subcaption}
\usepackage[normalem]{ulem}
\usepackage{hyperref}
\usepackage{booktabs}
\usepackage{color}
\usepackage{tabularray}

\usepackage{biblatex}
\bibliography{classic_algorithms_bibliography}

\title{Classic algorithms are fair learners: Classification Analysis of natural weather and wildfire occurrences}

\author{ \href{https://orcid.org/0000-0002-8144-1267}{\includegraphics[scale=0.06]{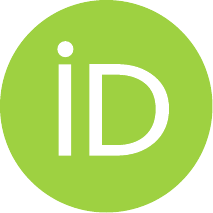}\hspace{1mm}Senthilkumar Gopal}\\
            College of Computing\\
            Georgia Institute of Technology\\
            \texttt{sengopal@gatech.edu}
}

\date{}
\renewcommand{\undertitle}{}
\renewcommand{\headeright}{}

\hypersetup{
pdftitle={Classic Algorithms are fair learners for limited data},
pdfsubject={cs.LG, cs.AI},
pdfauthor={Senthilkumar Gopal},
pdfkeywords={Machine Learning},
}

\begin{document}
\maketitle

\begin{abstract}
Classic machine learning algorithms have been reviewed and studied mathematically on its performance and properties in detail. This paper intends to review the empirical functioning of widely used classical supervised learning algorithms such as Decision Trees, Boosting, Support Vector Machines, k-nearest Neighbors and a shallow Artificial Neural Network. The paper evaluates these algorithms on a sparse tabular data for classification task and observes the effect on specific hyperparameters on these algorithms when the data is synthetically modified for higher noise. These perturbations were introduced to observe these algorithms on their efficiency in generalizing for sparse data and their utility of different parameters to improve classification accuracy. The paper intends to show that these classic algorithms are fair learners even for such limited data due to their inherent properties even for noisy and sparse datasets.

\end{abstract}

\keywords{Decision Trees \and Support Vector Machines \and k-nearest Neighbors \and Hyperparameter tuning}

\section{INTRODUCTION}

Though the classical Machine learning algorithms such as Decision trees, SVMs and k-nearest neighbors have been studied on their theoretical efficiency, practitioners and researchers tend to perform large scale hyper parameter searches or experimentation to identify the appropriate model and parameters everytime. There seems to be a need to do systematic study of these algorithms and these tuning parameters to quickly reduce the hypothesis space for identifying the best set of parameters to extract optimal performance. 

\textbf{Decision Trees: }Decision trees utilizes a tree structure to codify its learned strategy to classification the provided input. The ID3 algorithm ~\cite{quinlan1986induction} uses Entropy and Information gain to identify the nodes to split and arrive at the full tree during the training step. There are further improvements such as Boosting and pruning which can be tuned and investigated further to improve the accuracy of the tree.

\textbf{Support Vector Machines: }SVMs attempt to learn a decision boundary for linearly separable and non-linearly separable data (\textit{using kernel tricks}). The algorithm identifies the largest margin hyperplane that divides the data to perform classification and is effective even for higher dimension data.

\textbf{k-nearest Neighbors: }kNN learns the representation of the input data in the feature space using the nearest \textit{K }neighbors. This algorithm uses a lazy approach where there is no training time and the inference uses the nearest neighbors to determine the classification of new data points. However, this gets slower with a large number of samples or independent variables.

\textbf{Artificial Neural Network}: A Multi-layer perceptron classifier \cite{murtagh1991multilayer} is used to perform the classification task with varying hidden layer sizes as one of its hyperparameters. The MLP learns the layer weights as part of the training process to reduce the error using backpropagation.

All the code used to perform the experiments and results are published for reference
\footnote{https://github.com/sengopal/classic-ml-review-paper}

\section{RELATED WORK}

There has been multiple earlier works for analyzing classification ML algorithms on 112 real life binary datasets \cite{wainer2016comparison} to observe the functionalities of the classic algorithms. There has been previous in depth studies on tree based methods on the well documented UCI datasets \cite{zeng2022study} or non tabular datasets \cite{ates2021comparative} to analyze the performance of these algorithms on naturally occurring datasets with the intent on observing their effectiveness on these specific datasets. However, these explorations performed their analyzes with well rounded and natural datasets without any perturbations or synthesis for effective hyperparameter analysis, similar to \cite{zhang2017up}. Their findings around the effectiveness of gradient boosted trees over SVMs and decision trees were based on their documented datasets without any enquiry into how the hyperparameters would change their individual behavior. There has been some earlier work, \cite{holte1993very} to understand how levels of decision trees help, but this was only for decision trees and does not perform any active data perturbations and neither any extensive model parameter analysis.

\section{METHODOLOGY}

\subsection{Datasets}

The paper deters from using the commonly used UCI and other well established datasets to avoid running into ``\textit{statistical accidents}" as discussed in \cite{holte1993very}. To understand the algorithms and the effects that the hyper parameters have on them, the datasets need to be chosen producing relatively lower accuracy scores with the default algorithm implementations, but being responsive to various adjustments performed. The other criteria was to identify both a binary and a multi-class classification problem, to help understand the inherent behavior of the algorithms and use them to effectively evaluate them using various comparison metrics.

The first dataset - \textit{Rattle} \cite{noauthor_rain_nodate} represents the daily weather observations from Australian weather stations. This has around 56k samples and 65 features posing a non-trivial binary classification with tomorrow’s rain as the class of prediction.  The second dataset - Wildfire \cite{noauthor_1.88_nodate} contains data regarding the various US wildfires that occurred in the US from 1992 to 2015. However, this dataset has a relatively sparse feature set posing a completely different facet for investigating the algorithms.

Choosing these widely different datasets would help us explore the various supervised learning algorithms, their underpinnings, effects of their hyper parameters and how to tune for effective algorithmic performance. The two data sets used were particularly chosen for their sparsity and imbalance to study the effectiveness of these algorithms  The data has also been synthetically modified to help analyze the algorithms and garner their potential for classification problems where such data inefficiencies lie and identifies the hyperparameter tuning strategies which can be applied for other imbalanced and noisy data to optimize training for classification tasks.

\subsection{Preparation}

To understand the underpinnings of the algorithms and further analyze their model complexity, the datasets were reduced in size by certain filters. However, these reductions were carefully chosen to prevent any learner bias and caution was taken to avoid such biases by only choosing features with little to no correlation.

\begin{figure}[!htbp]
    \centering
    \subfloat[\centering Rain\_Tomorrow]{{\includegraphics[width=4.0cm]{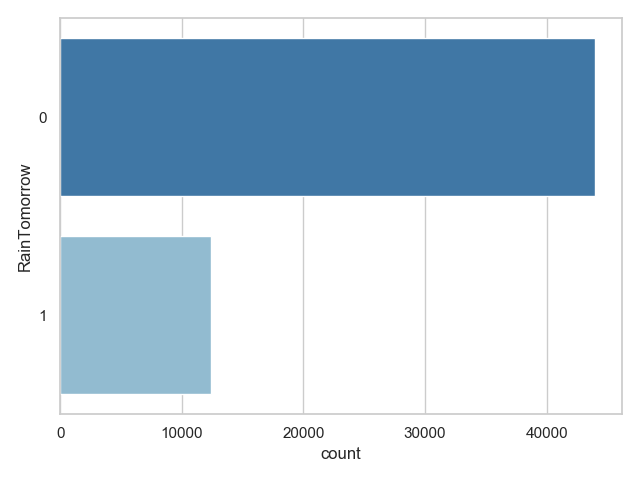} }}%
    \subfloat[\centering Wind\_Gust\_Direction]{{\includegraphics[width=4.0cm]{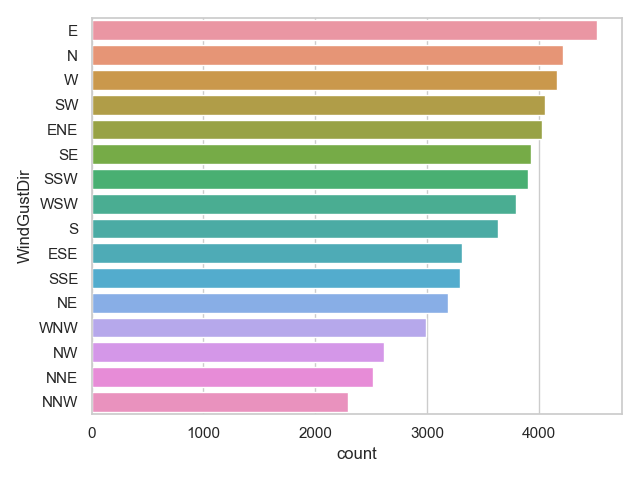} }}%
    \subfloat[\centering status\_cause\_code]{{\includegraphics[width=4.0cm]{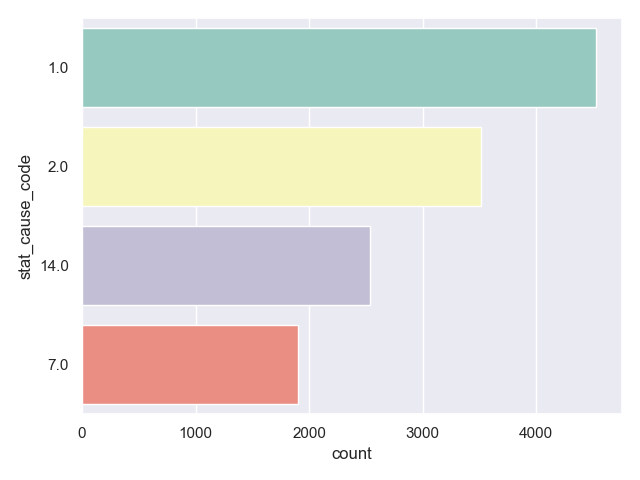} }}%
    \subfloat[\centering discovery\_month]{{\includegraphics[width=4.0cm]{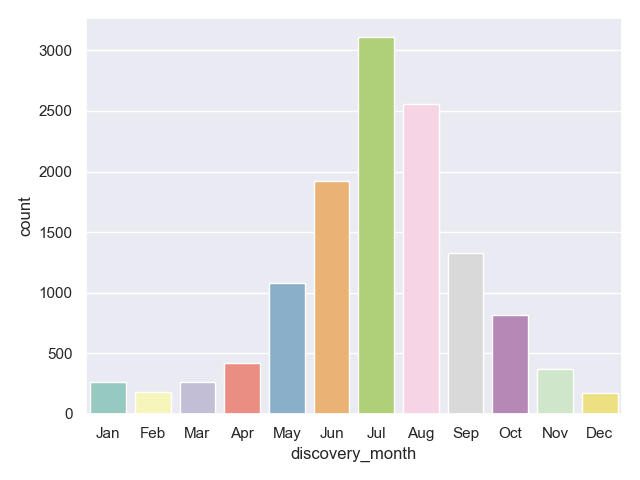} }}%
    \qquad
    \caption{Various features of the datasets - Rattle and Wildfire \label{fig:figure1}}%
\end{figure}

The Rattle dataset has 65 features and only the top 30 features were selected using the Chi-squared statistic to identify the most relevant ones. These features have the most impact on the outcome variable and hence the paper performs all the analyses using this reduced dataset. The distribution of some features are available at Figure \ref{fig:figure1} and the dataset characteristics are illustrated in Table \ref{table:table1}

\begin{table}[!htbp]
\centering
\begin{tabular}{|l|l|l|} 
\toprule
\textbf{Name} & \textbf{Rattle} & \textbf{Wildfire}  \\ 
\hline
Class         & Binary          & Multi-class        \\ 
\hline
Features      & 30              & 7                  \\ 
\hline
Sample size   & 56420           & 11825              \\ 
\hline
Balanced      & No              & Yes                \\
\bottomrule
\end{tabular}
\caption{Dataset characteristics \label{table:table1}}
\end{table}

The Wildfire dataset is filtered for only CA wildfires and the selection was based on domain knowledge rather than any statistical method. Also, feature data such as \textit{discovery\_date, cont\_date }have been modeled as categorical features by extracting the month and weekday of the year and the time taken for containment. \textbf{This dataset varies drastically from the previous one with larger categorical vs. continuous feature sets.} The dataset has been slightly modified to use only four classes of outcomes to create a more balanced dataset since some of the classes were very sparse and added the need to balance them which was beyond the scope of this paper.

\subsection{Intuition and Default results}

The results of using the algorithms with their default parameters are available in Table \ref{table:table2}. Along with the random classifier, they help form a baseline to tune the hyper parameters. These are simple accuracy metrics with no cross-validation and a 80/20 split for training and validation. Rattle dataset has a large sample size and predictably performs well across all algorithms. Due to the presence of more continuous variables, it performs slightly better while using ANN and SVM in comparison to Decision trees. While the Wildfire dataset with mostly categorical features fare better with Decision trees and with boosting in comparison to NN and SVM.  As noted by ~\cite{pal2003assessment}, the lower dimensionality wildfire dataset fared better with decision trees while the Rattle dataset with higher dimensionality performed better with neural networks and SVM.

\begin{table}[!htbp]
\centering
\renewcommand{\arraystretch}{1.3}
\begin{adjustbox}{max width=\textwidth}
\begin{tabular}{p{2.29cm}p{2.29cm}p{2.29cm}p{2.29cm}p{2.29cm}p{2.29cm}p{2.29cm}p{2.29cm}p{2.29cm}p{2.29cm}p{2.29cm}p{2.29cm}p{2.29cm}p{2.29cm}}
\hline
\multicolumn{1}{|p{2.29cm}}{\textbf{Algorithm}} & 
\multicolumn{1}{|p{2.29cm}}{Dummy Classifier} & 
\multicolumn{1}{|p{2.29cm}}{Decision Tree} & 
\multicolumn{1}{|p{2.29cm}}{Boosting} & 
\multicolumn{1}{|p{2.29cm}}{\textit{k-}Nearest Neighbors} & 
\multicolumn{1}{|p{2.29cm}}{Neural Network} & 
\multicolumn{1}{|p{2.29cm}|}{Support Vector Machines} \\ 
\hline
\multicolumn{1}{|p{2.29cm}}{\textbf{Rattle}} & 
\multicolumn{1}{|p{2.29cm}}{0.66} & 
\multicolumn{1}{|p{2.29cm}}{0.79} & 
\multicolumn{1}{|p{2.29cm}}{0.80} & 
\multicolumn{1}{|p{2.29cm}}{0.83} & 
\multicolumn{1}{|p{2.29cm}}{0.85} & 
\multicolumn{1}{|p{2.29cm}|}{0.85} \\ 
\hline
\multicolumn{1}{|p{2.29cm}}{\textbf{Wildfire}} & 
\multicolumn{1}{|p{2.29cm}}{0.27} & 
\multicolumn{1}{|p{2.29cm}}{0.47} & 
\multicolumn{1}{|p{2.29cm}}{0.55} & 
\multicolumn{1}{|p{2.29cm}}{0.51} & 
\multicolumn{1}{|p{2.29cm}}{0.44} & 
\multicolumn{1}{|p{2.29cm}|}{0.50} \\ 
\hline
\end{tabular}
\end{adjustbox}
\caption{Accuracy results of the algorithms using the default parameter values \label{table:table2}}
\end{table}

\subsection{Experiment Setup}

The following methodology and analysis discusses the effects of various parameters on all the five algorithms for these datasets and their respective change in accuracy metrics. 

\begin{figure}[!htbp]
    \centering
    \includegraphics[width=11.16cm,height=4.29cm]{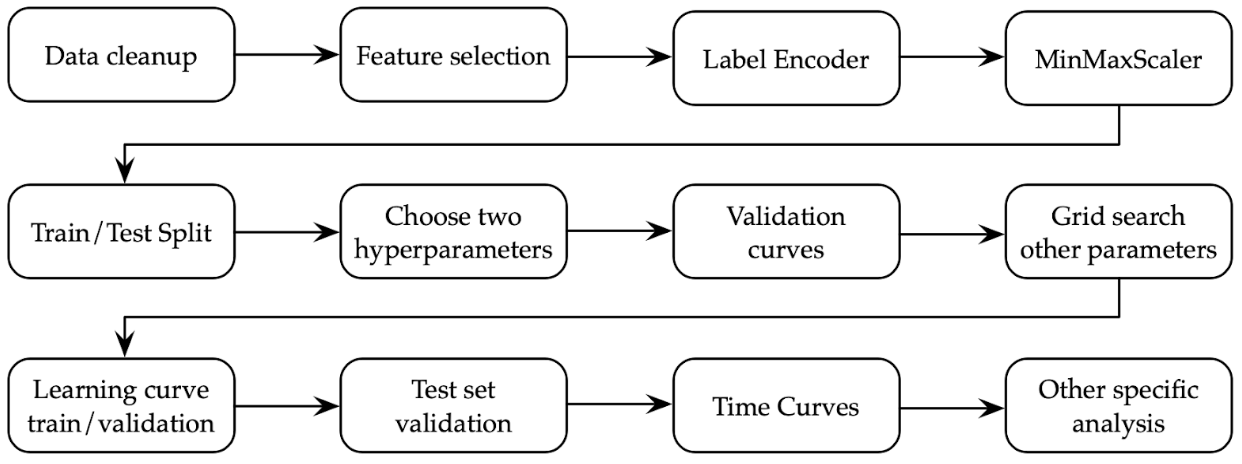}
    \caption{Various steps involved in the analysis process\label{fig:figure2}}%
\end{figure}

For the Rattle dataset, only the irrelevant data columns were removed and other columns were label encoded with min-max scaling for easier convergence and samples with unknown values were dropped. For the Wildfire dataset, the query constrained the dataset to have only samples with less unknown values and the multi-class data were merged and filtered to only 4 classes. There were also a few categorical features extracted using \textit{discovery\_time. }These were also label encoded and run through a min-max scaling as preparation.

\subsection{Model validation process flow}

The dataset is split initially broken into 80/20 split and the 20$\%$ split is used as the final hold out test data to verify the final tuned algorithms performance. All the training and cross validation happen on the 80$\%$ split which is further divided into train/validation sets for grid search, model complexity and learning curve analysis etc., while using a \textbf{3-fold }cross validation set. \textbf{Parameters which are not inherently embedded within the model as part of the learning from the dataset, but influence the functioning of the model are termed as hyper-parameters }and each algorithm is executed with a varying range of values for selected parameters. Grid search is employed to perform scoring across various combinations of the hyper parameter values to identify the optimal one. 

A learning curve analysis is then performed to determine various aspects of the model built such as bias, variance and how the model has generalized along with time curves. A final test of the optimized model is performed using the initial hold-out test data to measure the impact of improvements performed as part of the analysis. All these steps are performed in sequence as illustrated in Figure \ref{fig:figure2}.

\section{EVALUATION}

\subsection{Decision Trees}

A Decision Tree uses a tree-like model of decisions and consequences as a representation of the training data, and predicts outcomes for newer instances based on the tree model.

\subsubsection{Hyper parameter tuning}

\begin{figure}[!htbp]
\centering

\begin{subfigure}[b]{0.45\textwidth}
\centering
\includegraphics[width=\textwidth]{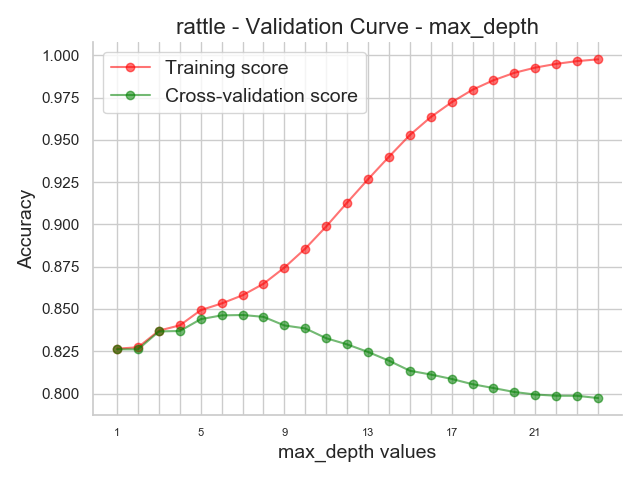}
\end{subfigure}
\hfill
   \begin{subfigure}[b]{0.45\textwidth}
\centering
\includegraphics[width=\textwidth]{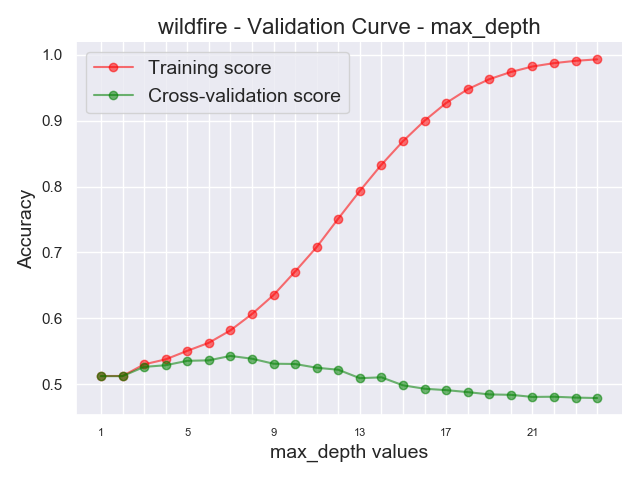}
\end{subfigure}
\caption{Effects of max\_depth parameter on training/validation accuracy using validation\_curve\label{fig:figure3}}
\end{figure}

\textbf{The \textit{max\_depth }parameter }determines the maximum depth of the tree. Due to its algorithmic nature,  decision trees are susceptible to overfitting leading to high variance with deeper trees which is very evident from \textit{max\_depth} graphs for both datasets. At low values of max\_depth, both datasets have high bias due to poor complexity, but around values 5-7, as per Figure~\ref{fig:figure3} the \textbf{training and validation curves start to diverge} indicating \textbf{overfitting and high variance at higher values of max\_depth}. 

\begin{figure}[!htbp]
\centering
\begin{subfigure}[b]{0.45\textwidth}
\centering
\includegraphics[width=\textwidth]{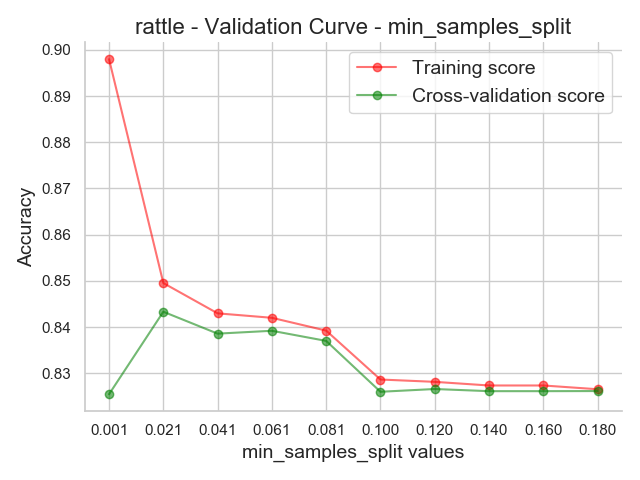}
\end{subfigure}
\hfill
 \begin{subfigure}[b]{0.45\textwidth}
\centering
\includegraphics[width=\textwidth]{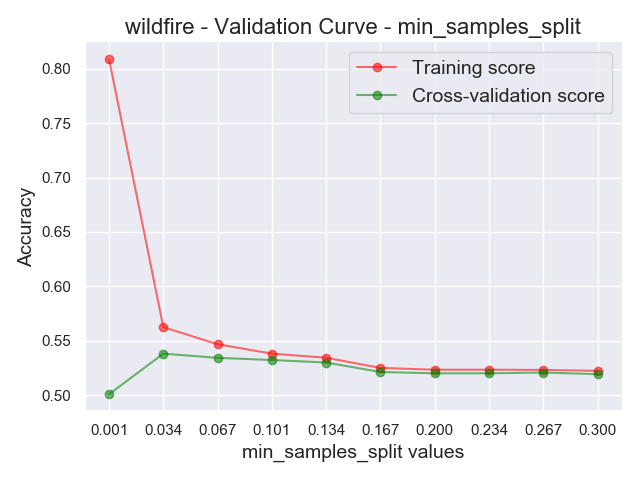}
\end{subfigure}
\caption{Effects of min\_samples\_split parameter on training/validation accuracy\label{fig:figure4}}
\end{figure}

\textbf{The \textit{min\_samples\_split }parameter} is the minimum number of samples required to split an internal node. The curve results exhibit that the model suffers from\textbf{ high bias and low variance, evident from high accuracy for low min\_samples\_splits where overfitting occurs}. However, since the validation score holds steady, the model conforms to low variance and would need tuning to balance the high bias. Using the plots referred in Figure~\ref{fig:figure4}, a range of values for \textit{max\_depth} and \textit{min\_samples\_split } were determined and using a gridsearch, the optimal combination was identified.

As Balakrishnan \cite{balakrishnanbits} states, entropy is defined ''as the average or expected uncertainty associated with a set of events'' and computes using the log base 2 of probabilities in comparison to Gini index which calculates only on simple probabilities and works usually well for continuous feature variables. The criteria for attribute split was determined as \textbf{entropy} instead of gini index due to the presence of more categorical attributes and entropy usually performs better in such models. This was verified using grid search as well.

\subsubsection{Model Complexity Analysis}

\begin{figure}[!htbp]
\centering
\begin{subfigure}[b]{0.45\textwidth}
\centering
\includegraphics[width=\textwidth]{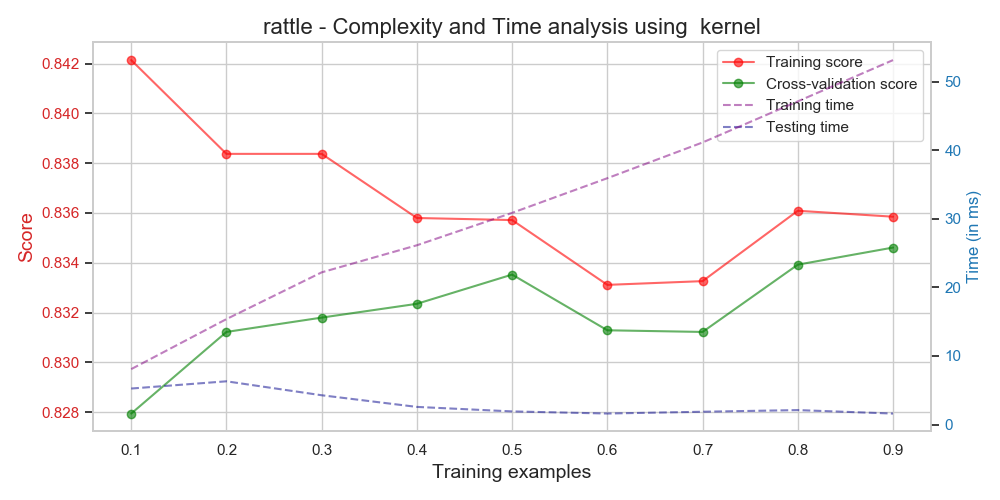}
\end{subfigure}
\hfill
\begin{subfigure}[b]{0.45\textwidth}
\centering
\includegraphics[width=\textwidth]{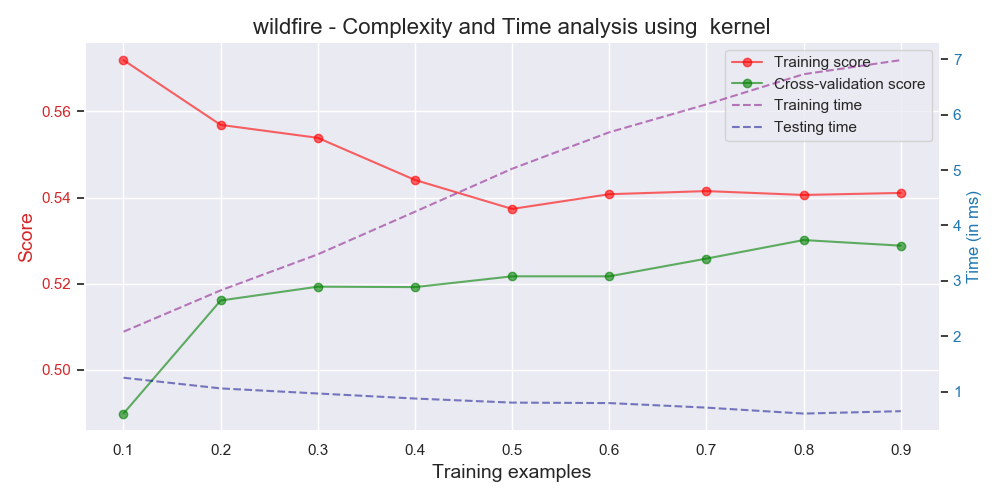}
\end{subfigure}
\caption{Learning Curve Model Complexity Analysis for Decision Trees\label{fig:figure5}}
\end{figure}

Once the optimal hyper parameters are identified, model complexity analysis can be performed to identify the bias-variance trade-offs attributed in the model. Plotting the learning curve in Figure~\ref{fig:figure5} indicates that for the Rattle dataset, the curves have converged indicating low variance. However, the training score starts high and reduces with training size, indicating the model has trouble identifying more variations of the training data size i.e., \textbf{high bias}. Since the curves have converged and due to high bias rather than adding more sample data, \textbf{additional features are required to improve the model’s complexity}, as the model seems to be slightly overfit.

For the Wildfire dataset, a similar outcome is identified. The model has\textbf{ higher bias and lower variance}. As the curves have not converged, they can definitely gain from more sample data to help reduce variance and also need more features for improving complexity and reducing the bias. With increasing training sample size, the training time also increases linearly as the tree is being built during training, while the prediction time is constant as it involves a simple lookup from the earlier built tree.

\subsubsection{Pruning and its effects}

\definecolor{Silver}{rgb}{0.8,0.8,0.8}
\begin{table}[!htbp]
\centering
\begin{tblr}{
  column{3} = {Silver},
  column{5} = {Silver},
  cell{1}{2} = {c=2}{},
  cell{1}{4} = {c=2}{},
  hlines,
  vlines,
}
          & \textbf{Rattle}  &                 & \textbf{Wildfire} &                 \\
          & \textbf{Current} & \textbf{Pruned} & \textbf{Current}  & \textbf{Pruned} \\
Accuracy  & 0.83             & 0.84            & 0.55              & 0.53            \\
Precision & 0.82             & 0.82            & 0.52              & 0.48            \\
Recall    & 0.83             & 0.84            & 0.55              & 0.53            \\
Branches  & 13               & 6               & 31                & 15              \\
Nodes     & 27               & 13              & 63                & 31              
\end{tblr}
\caption{Results with and without pre-pruning \label{table:table3}}
\end{table}

\begin{figure}[!htbp]
\centering
\begin{subfigure}[b]{0.45\textwidth}
\centering
\includegraphics[width=\textwidth]{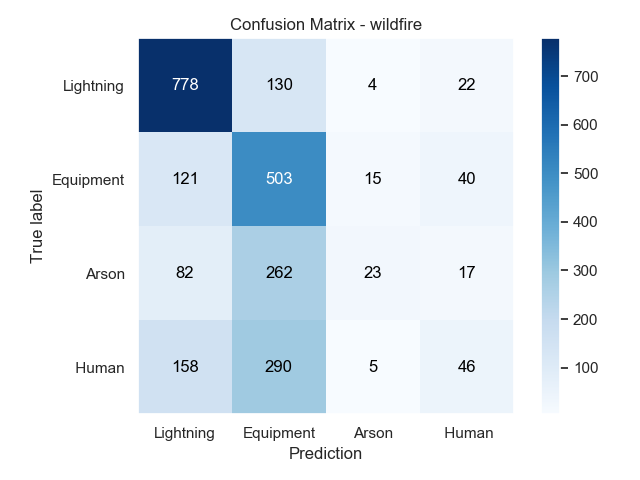}
\end{subfigure}
\hfill
\begin{subfigure}[b]{0.45\textwidth}
\centering
\includegraphics[width=\textwidth]{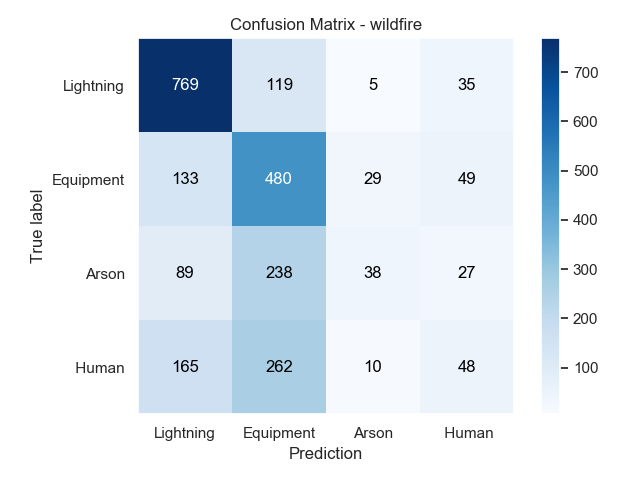}
\end{subfigure}
\caption{Confusion matrices before and after pruning for Wildfire\label{fig:figure6}}
\end{figure}

Pruning of decision trees helps reduce the tree size by eliminating parts of the tree that contribute very little to the classification problem. Pruning addresses the over fitting problem by reducing bias and eliminating complexity. Pruning can be achieved by using either early stopping/pre-pruning or post-pruning techniques. The following pre-pruning techniques have been utilized successfully with minimal to no loss of accuracy and also achieving better generalization - Reducing the depth of the tree (using \textit{max\_depth} parameter) and using \textit{min\_samples\_leaf }to force splits, only when there are a minimum number of samples available in that node and the results are available in Table~\ref{table:table3} and the corresponding confusion matrices are available in Figure~\ref{fig:figure6}

Another interesting aspect of this analysis, shows the contrast and trade-off between information gain (achieved using gini/entropy) versus the accuracy metric which determines the growth and pruning of the tree respectively.

\begin{figure}[!htbp]
\centering
\begin{subfigure}[b]{0.45\textwidth}
\centering
\includegraphics[width=\textwidth]{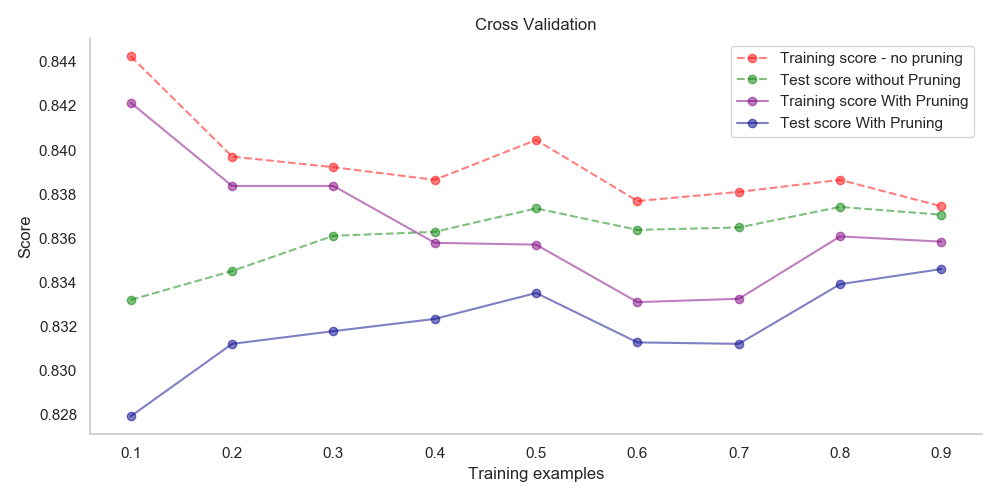}
\end{subfigure}
\hfill
\begin{subfigure}[b]{0.45\textwidth}
\centering
\includegraphics[width=\textwidth]{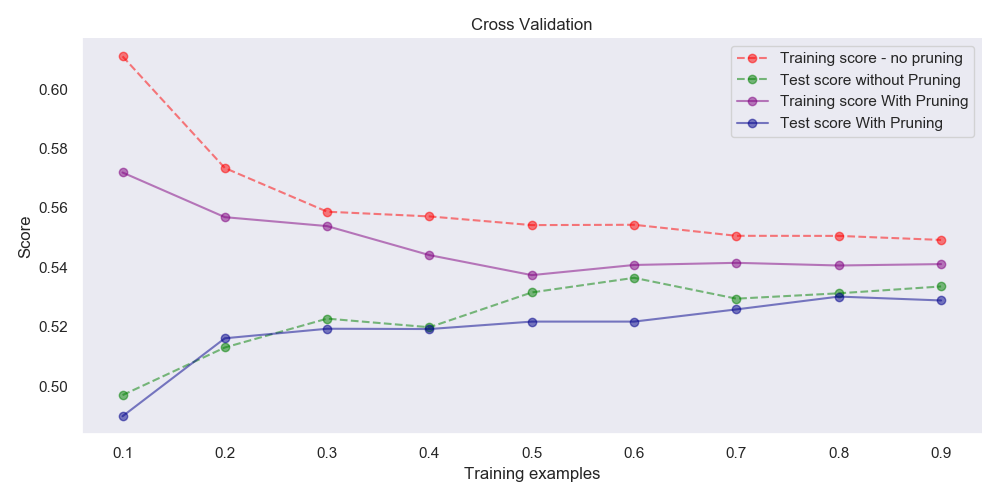}
\end{subfigure}
\caption{Learning curves of Rattle and Wildfire with scores for current and pruned Decision Trees\label{fig:figure7}}
\end{figure}

On plotting the learning curves with and without pruning, we can see that pruning helps generalization better and helps achieve better accuracy over validation set as evident in Figure~\ref{fig:figure7}. We can also notice \textbf{reduced bias (\textit{lower training scores at lower samples})} and \textbf{reduced variance} as well. The effect is more prominent in the Wildfire dataset as the training scores have gone down significantly for lower training sizes producing a lower bias strain.

\subsection{Boosting}

Boosting is an ensemble algorithm aimed at reducing bias and variance by using a family of weak learners to create a strong classifier. A weak learner is simply a classifier that is very lightly correlated with actual classification, while a strong learner is well-correlated with the classification, such as the decision tree classifier built earlier.

\subsubsection{Hyper parameter tuning}

Along with tuning the parameters, the analysis also observes the effectiveness of boosting while using the pruned and default decision trees.

\begin{figure}[!htbp]
\centering
\begin{subfigure}[b]{0.45\textwidth}
\centering
\includegraphics[width=\textwidth]{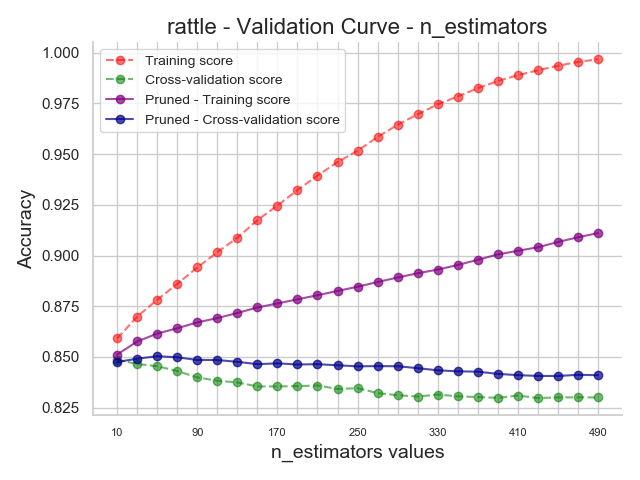}
\end{subfigure}
\hfill
\begin{subfigure}[b]{0.45\textwidth}
\centering
\includegraphics[width=\textwidth]{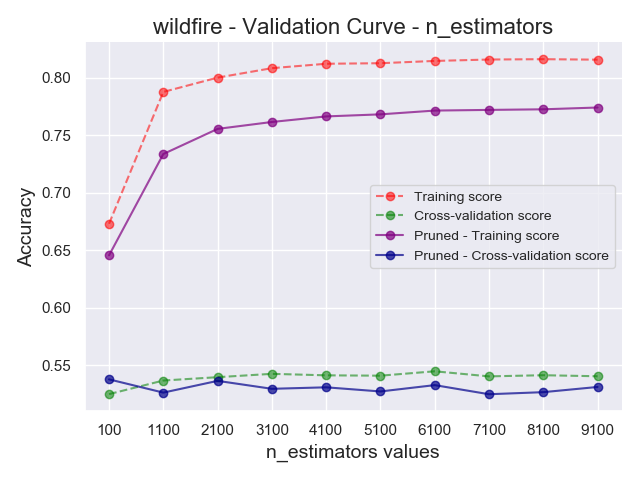}
\end{subfigure}
\caption{Effects of n\_estimators parameter on accuracy for both non-pruned $\&$ pruned trees as estimators\label{fig:figure8}}
\end{figure}

The \textbf{\textit{n\_estimators }parameter} indicates the maximum number of estimators to use for boosting. The effects of boosting on the accuracy using the unpruned and pruned trees display an interesting behavior where the unpruned tree being a strong classifier, starts to overfit right from the beginning even with a low number of estimators. However, the well generalized pruned tree improves accuracy with boosting with the addition of more estimators before the overfitting begins. Theortically boosting helps generalize better, but however the datasets that have been chosen seem to be resistant, which might be due to noise/misclassification or lack of features in case of Wildfire dataset leading to \textbf{low bias and high variance} as illustrated in Figure~\ref{fig:figure8}

Boosting with multiple classifiers helps address the high bias problem that was earlier observed. However, due to the lack of features and enough sample data, high variance can be addressed only to a certain extent.

\begin{figure}[!htbp]
\centering
\begin{subfigure}[b]{0.45\textwidth}
\centering
\includegraphics[width=\textwidth]{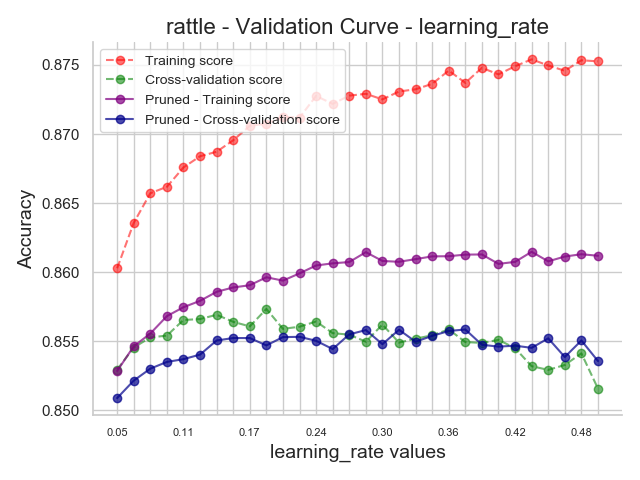}
\end{subfigure}
\hfill
\begin{subfigure}[b]{0.45\textwidth}
\centering
\includegraphics[width=\textwidth]{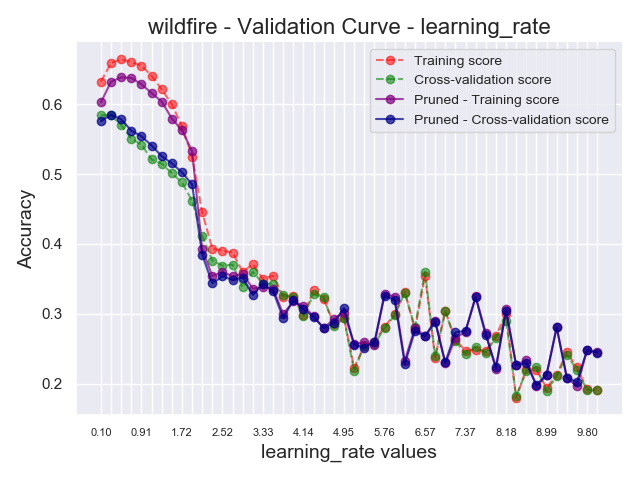}
\end{subfigure}
\caption{Effects of learning\_rate parameter on accuracy for both non-pruned $\&$ pruned trees as estimators\label{fig:figure9}}
\end{figure}

\textbf{The \textit{learning\_rate} parameter }determines the contribution of each classifier and is usually used to offset large estimators with a lower learning rate to help generalize. The \textit{learning\_rate }accuracy vs. training size using both the unpruned and pruned trees paints an interesting picture. With higher \textbf{\textit{learning\_rate }}values, the classifiers get more weightage for their decisions and with lesser classifiers contributing more, overfitting can be observed when the learning\_rate increases, leading to the theory that \textbf{high bias cannot be addressed using learning\_rate, though low variance }seems to be obtained as seen in Figure~\ref{fig:figure9}

In both datasets, the interesting aspect is the resistance offered by the pruned tree for overfitting until larger values of \textit{n\_estimators} and \textit{learning\_rate, }signifying their importance while using Decision trees.

\subsubsection{Model Complexity Analysis}

\begin{figure}[!htbp]
\centering
\begin{subfigure}[b]{0.45\textwidth}
\centering
\includegraphics[width=\textwidth]{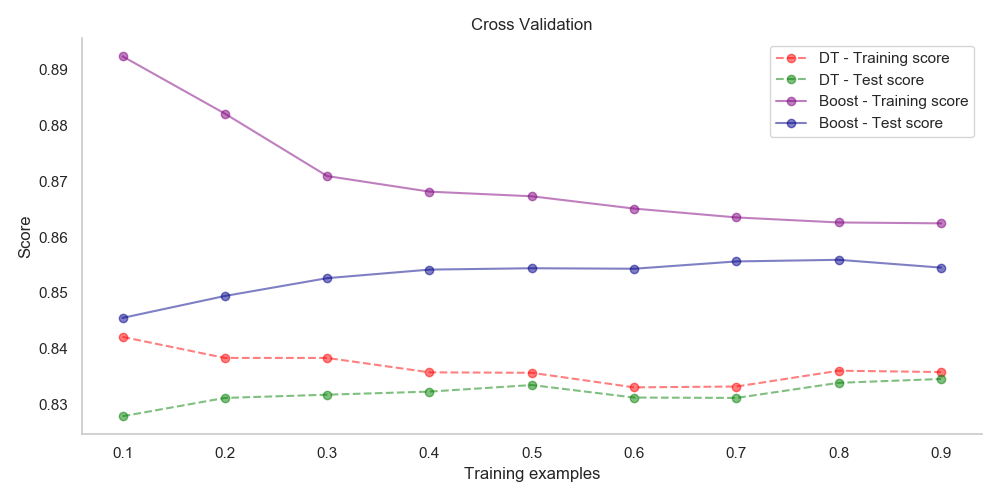}
\end{subfigure}
\hfill
\begin{subfigure}[b]{0.45\textwidth}
\centering
\includegraphics[width=\textwidth]{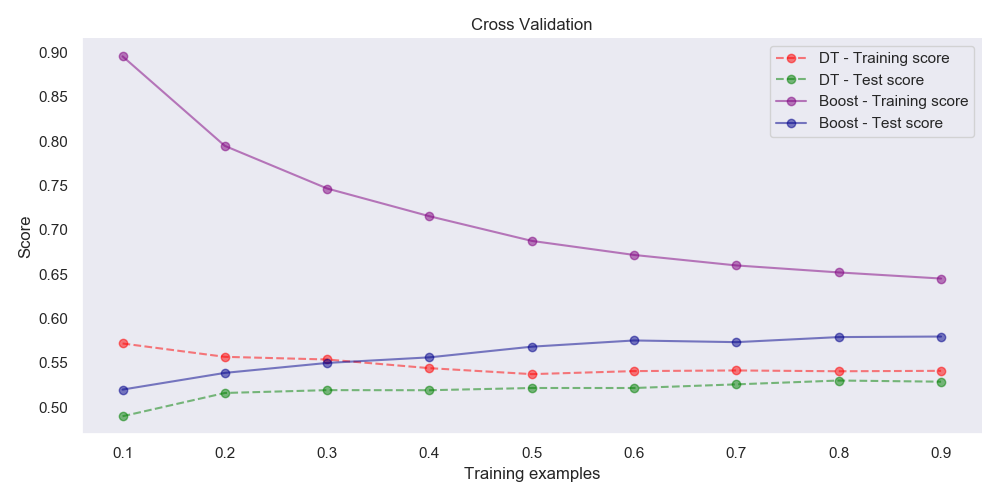}
\end{subfigure}
\caption{Learning curves for Boosting Algorithm in comparison with Decision tree\label{fig:figure10}}
\end{figure}

Plotting the learning curve using the optimal hyper parameter values, we can observe that boosting has not completely addressed the high bias as shown in Figure~\ref{fig:figure10}. Training error is still high for lower sampling values indicating the \textbf{fallacy of using Boosting with a relatively strong learner}. As observed, the models with boosting have better accuracy values, but still have \textbf{high bias and similar variance in comparison to decision tree algorithms}. We can also observe that the testing accuracy increases better in comparison to the decision tree learning curve, allowing the training and testing accuracies to converge, indicating low variance and enabling addition of more sample data to help reduce variance. The effect of boosting is observed in the Wildfire dataset as well, with a \textbf{higher bias} due to its lack of features which gets amplified in Boosting.

\begin{figure}[!htbp]
\centering
\begin{subfigure}[b]{0.45\textwidth}
\centering
\includegraphics[width=\textwidth]{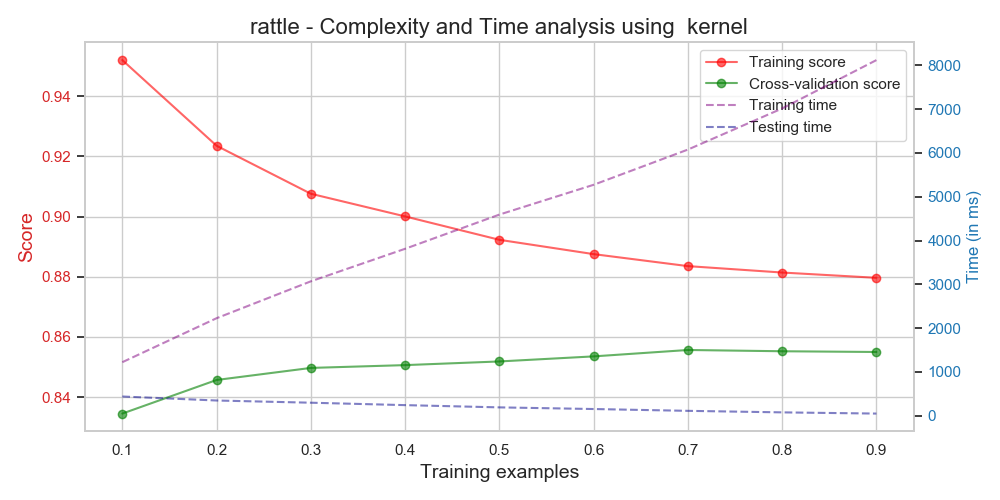}
\end{subfigure}
\hfill
\begin{subfigure}[b]{0.45\textwidth}
\centering
\includegraphics[width=\textwidth]{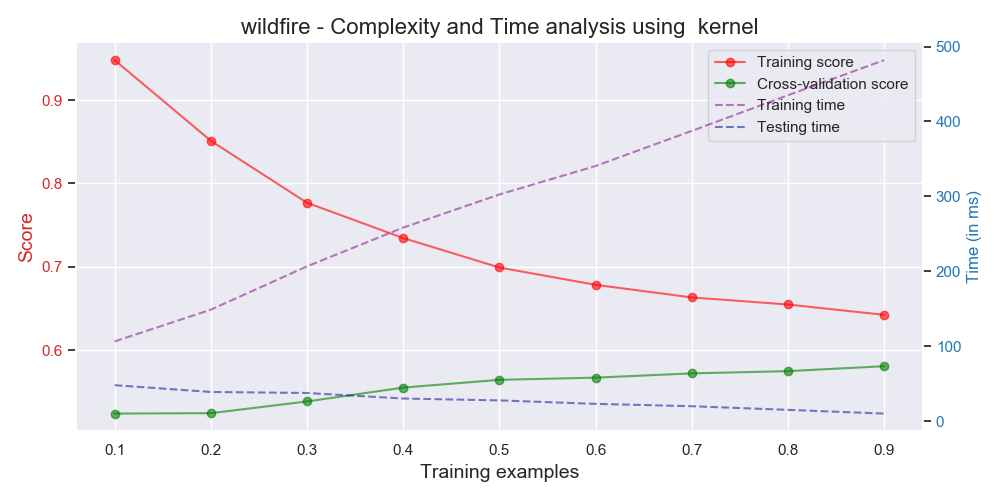}
\end{subfigure}
\caption{Learning curves for Boosting Algorithm capturing time for training and test\label{fig:figure11}}
\end{figure}

A key observation is the effectiveness of Boosting while using a relatively strong learner as base estimator. As the underlying learner is not a weak learner (\textit{closer to random results}), the effectiveness of Boosting is not too high in comparison with the underlying base learner with a mere 3-4$\%$ increase in accuracy. However, it is also worthy to note that the underlying classifier already has high accuracy values (Rattle) dataset and Boosting still was able to increase its high accuracy further. Also, boosting hyper parameters cannot completely eradicate overfitting, if the underlying learner is already strong and overfits to some extent. As expected, Boosting helps reduce the error boundaries, however fails to address the bias in the model. Similar to Decision trees, the training time increases linearly with the sample size, while the prediction time is constant as well described in Figure~\ref{fig:figure11}

\textit{Note: Boosting rarely overfits for low variance datasets and can be tuned with very few hyperparameters in comparison to XGBoost~\cite{chen2016xgboost} which is finetuned for speed and performance with a multitude of hyperparameters and is beyond the range of this experiment.}

\subsection{k-Nearest Neighbors}

k-Nearest Neighbors classification performs instance-based learning instead of deriving an internal model, by saving the training data samples. Prediction is computed using the majority vote of nearest neighbors determining the class having most representation among the identified neighbors.

\subsubsection{Hyper parameter tuning}

\textbf{The n\_neighbors parameter} is the critical parameter for kNN as it determines the number of neighbors to consider for classifying the request sample. An interesting observation was the \textbf{high bias} (large training accuracy and small testing accuracy) observed when the neighbors were weighted using ``\textbf{\textit{distance}}" instead of ``\textbf{\textit{uniform}}" weights. Intuition suggests this might be due to the presence of a large set of samples within small ranges which influences the result a lot more due to its closeness. To avoid this overfitting only ``\textbf{\textit{uniform}}" weights were used.

\begin{figure}[!htbp]
\centering
\begin{subfigure}[b]{0.23\textwidth}
\centering
\includegraphics[width=\textwidth]{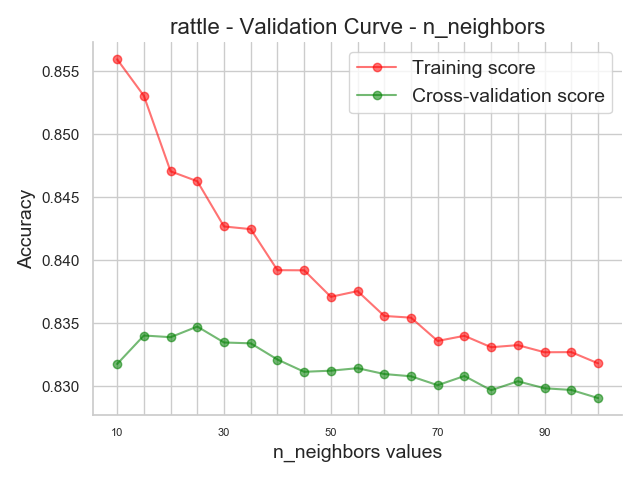}
\end{subfigure}
\hfill
\begin{subfigure}[b]{0.23\textwidth}
\centering
\includegraphics[width=\textwidth]{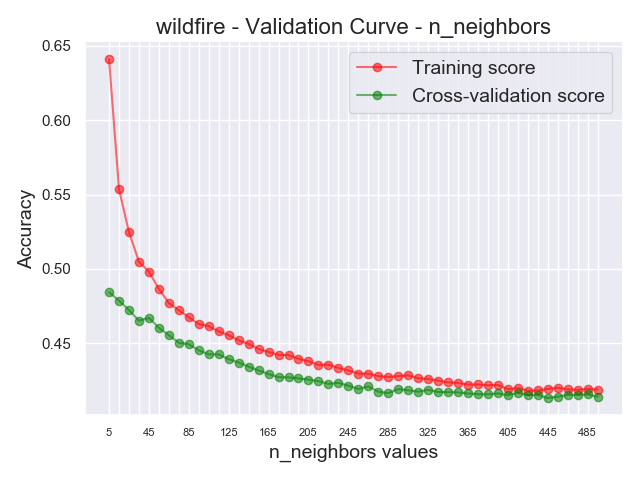}
\end{subfigure}
\hfill
 \begin{subfigure}[b]{0.23\textwidth}
\centering
\includegraphics[width=\textwidth]{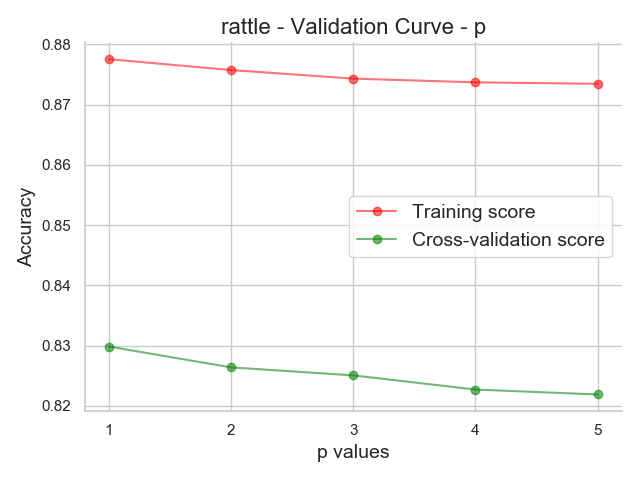}
\end{subfigure}
\hfill
\begin{subfigure}[b]{0.23\textwidth}
\centering
\includegraphics[width=\textwidth]{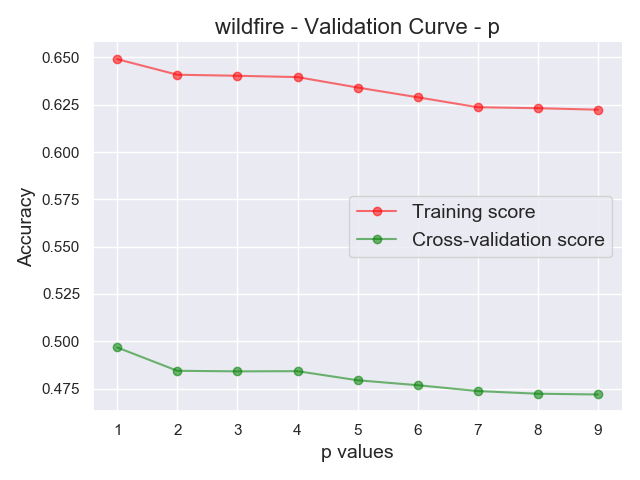}
\end{subfigure}
\caption{Effects of \textit{n\_neighbors }and the \textit{p} parameter on training/validation accuracy\label{fig:figure12}}
\end{figure}

The complexity analysis presents an expected phenomenon where with k$=$1 or low values\textbf{, the training data is overfit with high bias}. However, with increasing neighbors we were able to observe convergence with the testing accuracy curve. Surprisingly in comparison to earlier algorithms, the testing score remains fairly steady for Rattle and with slight changes for Wildfire indicating a \textbf{low variance model} with a larger neighbor count, but reinforces the \textbf{high bias hypothesis}.

\textbf{The p (distance metric) parameter} is another parameter that was analyzed. \textit{Chomboon, Kittipong, et al }mentions that Euclidean, Manhattan and Minkowski have similar accuracy metrics and performing complexity analysis produces an interesting data point where Euclidean distance outperforms the rest by a minor margin. All the \textit{p }values exhibited \textbf{high variance with low bias, }drawing the conclusion that they do not contribute to reducing any model overfitting, rather providing incremental improvements for accuracy.

The effects of various values for both these parameters are illustrated in Figure~\ref{fig:figure12}

\subsubsection{Model Complexity Analysis}
\begin{figure}[!htbp]
\centering
\begin{subfigure}[b]{0.45\textwidth}
\centering
\includegraphics[width=\textwidth]{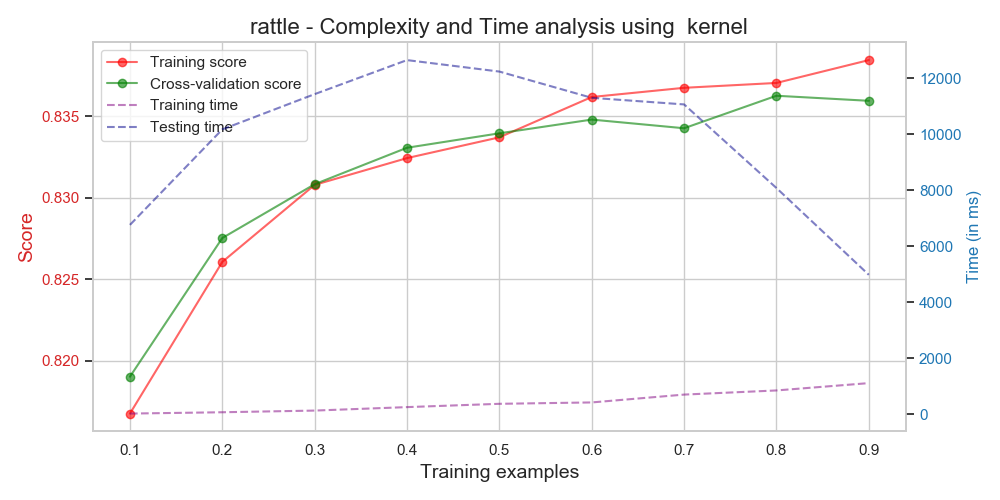}
\end{subfigure}
\hfill
\begin{subfigure}[b]{0.45\textwidth}
\centering
\includegraphics[width=\textwidth]{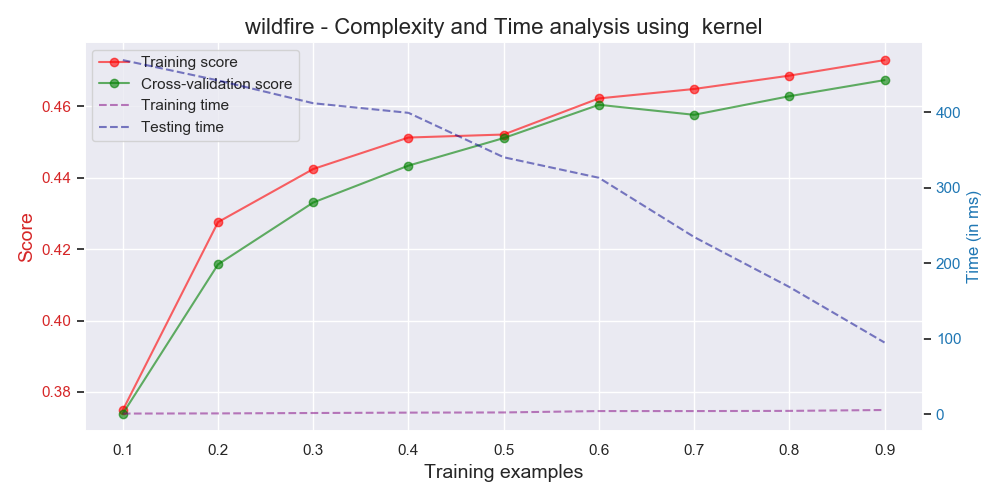}
\end{subfigure}
\caption{Learning curves for kNN for complexity analysis along with time for training and test\label{fig:figure13}}
\end{figure}

Plotting the learning curve using the optimal parameters identified, as in Figure~\ref{fig:figure13}, produces some telling results. Using the appropriate number of neighbors helps generalize the model even for low training samples. The accuracy is slightly low for less training samples, however the model displays \textbf{low bias} and \textbf{low variance} and continues the trend across different sample sizes. Increasing the number of neighbors would probably lower the variance with a penalty for accuracy or more training samples can be added to help improve accuracy and reduce variance.

The same phenomenon was observed in both datasets indicating kNN as a good meta-classification algorithm for these types of datasets. Due to its nature of merely saving the sample instead of generating any internal representations, the training time is low and constant, while the prediction time decreases with increase of training sample set.

\subsection{Artificial Neural Networks (ANN)}

Artificial Neural networks (ANN) are based on brain-like systems of input, output layers and hidden layers similar to neurons to process data and produce learning based outputs. They use multiple layers and weighted perceptrons to ``learn" and represent the model of learning a particular problem dataset. A multilayer perceptron (MLP) is one such ANN where each node uses a nonlinear activation function.

\subsubsection{Hyper parameter tuning}

Due to the wide variety of hyper parameters being available for tuning, few of these parameters were verified for their usability and disregarded based on their accuracy results. For instance, the \textbf{\textit{‘adam’ solver}} was chosen as per sklearn recommendation for large datasets and a brief accuracy comparison with other solvers. \textbf{\textit{momentum}} parameter is utilized only with ‘sgd’ solver and did not yield any accuracy contributions and was disregarded.

\begin{figure}[!htbp]
\centering
\begin{subfigure}[b]{0.45\textwidth}
\centering
\includegraphics[width=\textwidth]{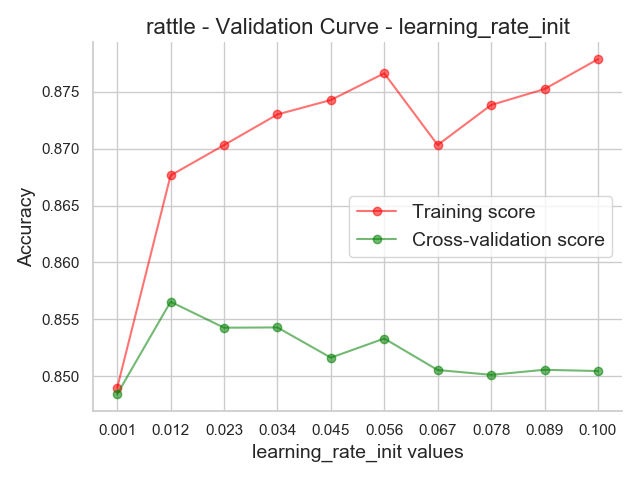}
\end{subfigure}
\hfill
 \begin{subfigure}[b]{0.45\textwidth}
\centering
\includegraphics[width=\textwidth]{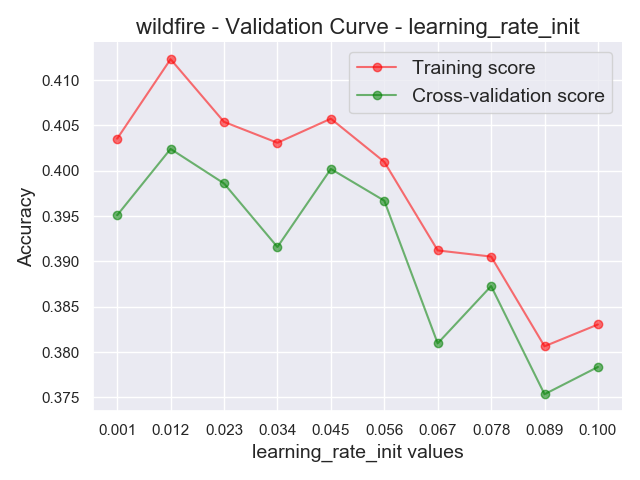}
\end{subfigure}
\caption{Effects of \textit{learning\_rate\_init} parameter on training/validation accuracy\label{fig:figure14}}
\end{figure}

\textbf{The \textit{learning\_rate\_init} parameter} yields an interesting curve with \textbf{low bias and high variance} for the Rattle dataset where there are a large number of features and a larger dataset, while offering a\textbf{ high bias and low variance} model for the sparsely featured Wildfire dataset, as shown in Figure~\ref{fig:figure14}

\textbf{The \textit{alpha} parameter }does not seem to offer any help in addressing bias and seems to indicate a \textbf{low bias }since the training score does not decrease with increase in sample size. However, the gap does indicate a \textbf{high variance} which can be addressed only by adding more data or possibly increasing the learning\_rate or performing re-sampling of data to generate newer data. A similar phenomenon is observed with the Wildfire data set as well.

\textbf{The \textit{hidden\_layers} parameter }is not really a comparable evaluation parameter using a graph. However, the graphs were generated with increasing order of complexity of hidden layers to analyze their performance. As expected the accuracy scores do increase with an increase in hidden layers. However, there is little to no indication of bias which seems to make ANN an attractive model for these datasets. There are slight indications of overfitting in the Rattle dataset while using 5-layers of 30 nodes. However, the wildfire dataset seems stable with \textbf{low bias and low variance}. Some more sample data might help address the slight variance, but the model requires more features to improve its complexity and accuracy scores.

\begin{figure}[!htbp]
\centering
\begin{subfigure}[b]{0.23\textwidth}
\centering
\includegraphics[width=\textwidth]{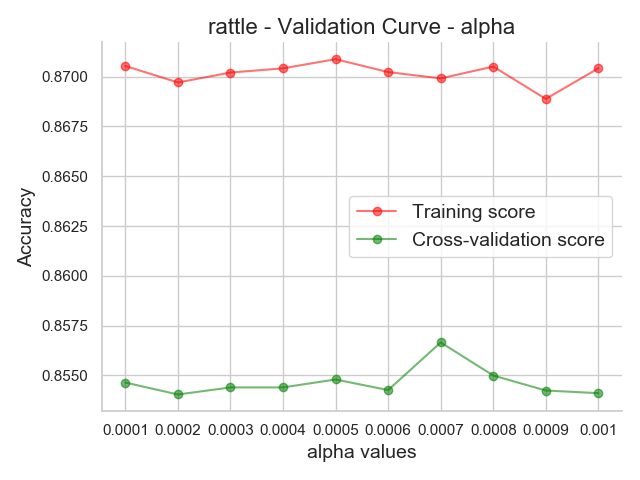}
\end{subfigure}
\hfill
\begin{subfigure}[b]{0.23\textwidth}
\centering
\includegraphics[width=\textwidth]{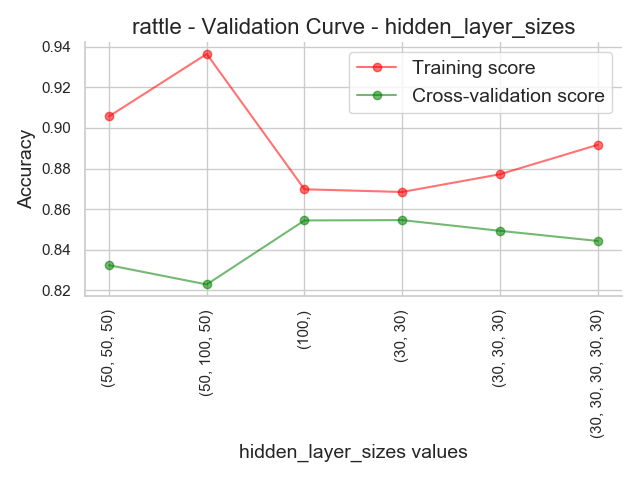}
\end{subfigure}
\hfill
\begin{subfigure}[b]{0.23\textwidth}
\centering
\includegraphics[width=\textwidth]{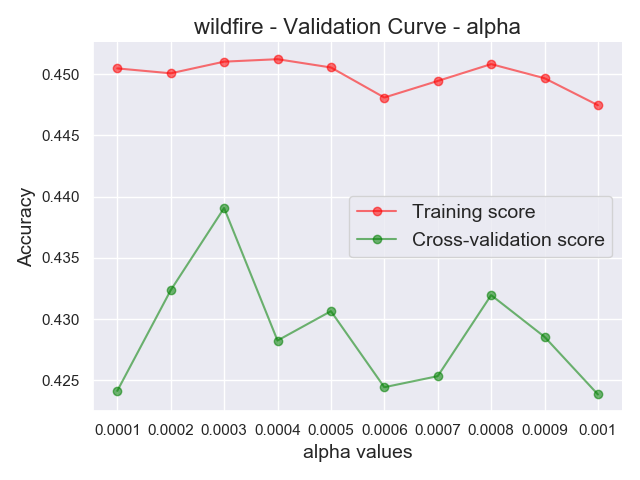}
\end{subfigure}
\hfill
\begin{subfigure}[b]{0.23\textwidth}
\centering
\includegraphics[width=\textwidth]{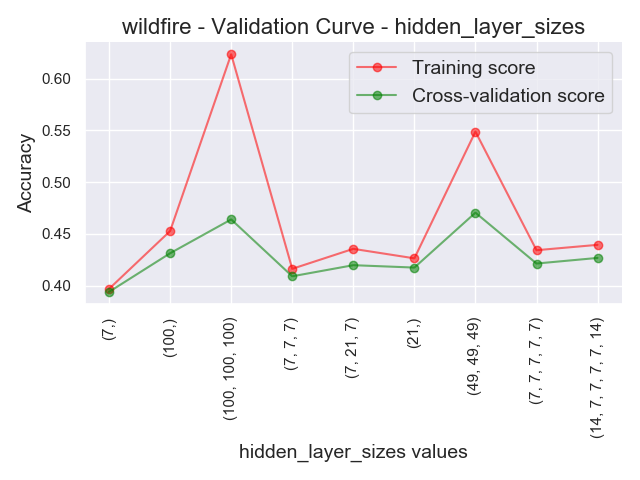}
\end{subfigure}
\caption{Effects of \textit{alpha and hidden\_layer\_sizes} parameters on training/validation accuracy\label{fig:figure15}}
\end{figure}

The effects of various values for both parameters are shown in detail in Figure~\ref{fig:figure15}.

\subsubsection{Model Complexity Analysis}
Both the datasets have mostly categorical features and rattle has 30 features lending itself to be solved by the sigmoid activation function. Though they have slow convergence, the presence of a rich feature set and a shallow network of (30,30) makes \textbf{sigmoid} the best suited activation for Rattle dataset. However, due to the sparsity of features and to prevent the vanishing gradient problem in sigmoid activation, the \textbf{tanh} activation was used instead for the Wildfire dataset.

\begin{figure}[!htbp]
\centering
\begin{subfigure}[b]{0.45\textwidth}
\centering
\includegraphics[width=\textwidth]{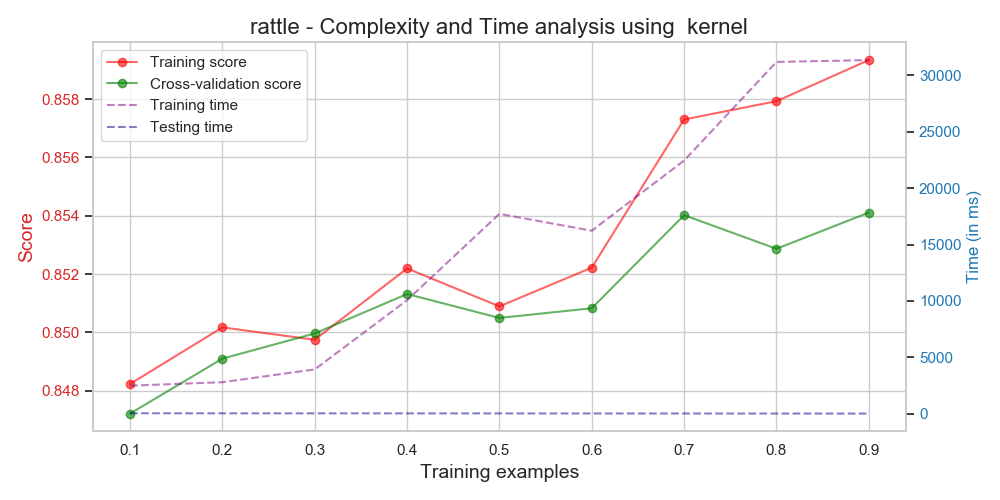}
\end{subfigure}
\hfill
\begin{subfigure}[b]{0.45\textwidth}
\centering
\includegraphics[width=\textwidth]{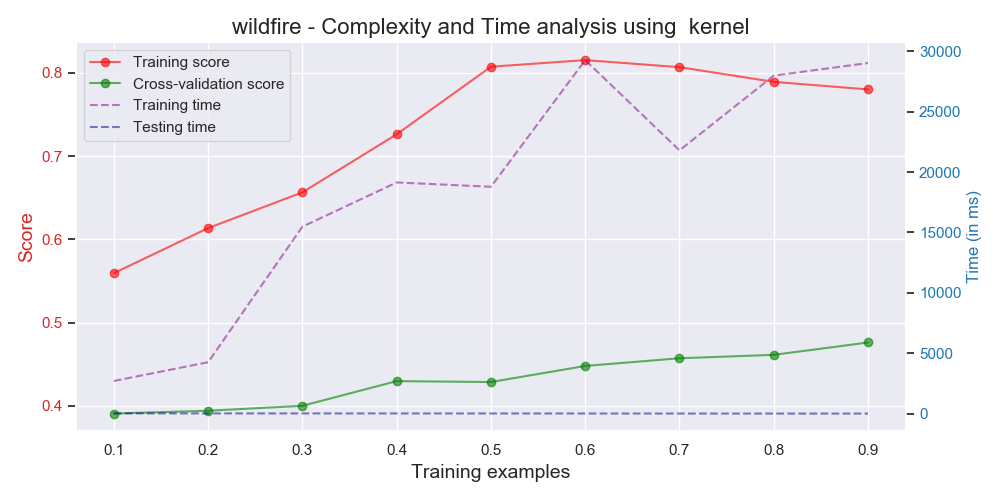}
\end{subfigure}
\caption{Learning curves for ANN for complexity analysis across training sample size\label{fig:figure16}}
\end{figure}

Relu activation was considered, but grid search proved sigmoid and tanh better for these particular datasets and intuitively they function better for classification problems than Relu. The hidden layer network was modeled intuitively with one node per feature (or its multiples) and \textit{sequential orthogonal approach ~\cite{panchal2014review}} where one layer after another is added for error minimization.

\begin{figure}[!htbp]
\centering
\begin{subfigure}[b]{0.45\textwidth}
\centering
\includegraphics[width=\textwidth]{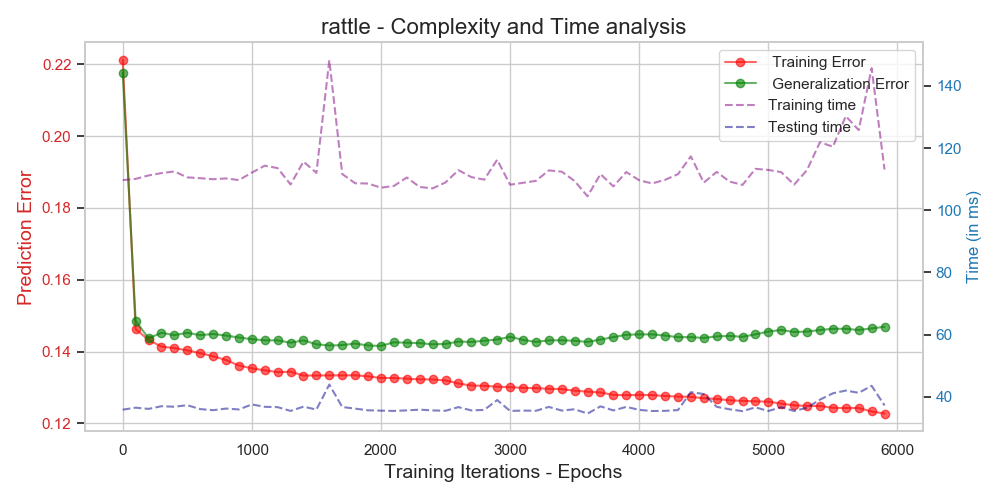}
\end{subfigure}
\hfill
\begin{subfigure}[b]{0.45\textwidth}
\centering
\includegraphics[width=\textwidth]{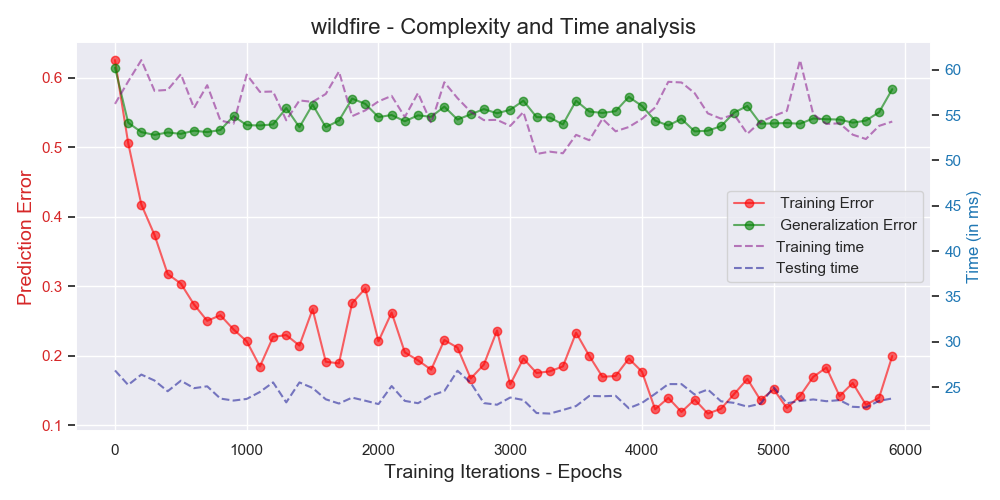}
\end{subfigure}
\caption{Learning curves for ANN for varied number of iterations/epochs\label{fig:figure17}}
\textit{Note: This learning curve is plotted with error metric in y-axis while others are plotted with accuracy}
\end{figure}

Plotting the error rate against the number of iterations produces an interesting observation where the time for training and testing remains constant across the number of iterations. However, with increase in iterations count, the feature rich Rattle dataset, starts to overfit slightly (\textbf{high bias and low variance}) while the sparsely featured Wildfire dataset overfits pretty quickly. It is evident that the model for Wildfire suffers from\textbf{ high bias} and \textbf{high variance} and overfitting which might be due to the larger number of nodes in each hidden layer as shown in Figure~\ref{fig:figure16} and Figure~\ref{fig:figure17}

The Rattle dataset might fare better from slight adjustments to the number of nodes and adding more layers to reduce variance and using regularization methods such \textbf{early stopping }and \textbf{L1 regularization} to reduce the number of features.\textbf{ }While for the Wildfire dataset, the number of nodes in each layer needs to be reduced using \textbf{drop-out regularization} and by adding more features to improve the model's complexity.

\subsection{Support Vector Machines}

SVM trains a model functioning as a non-probabilistic linear classifier using sample representation of spatial coordinates and plotted in such a manner to achieve clear separability across classes. For this analysis, the C- Support Vector Classifier (SVC) with different kernels is utilized instead of a Stochastic Gradient Descent which is usually recommended for larger datasets.

\subsubsection{Hyper parameter tuning}

\begin{figure}[!htbp]
\centering
\begin{subfigure}[b]{0.45\textwidth}
\centering
\includegraphics[width=\textwidth]{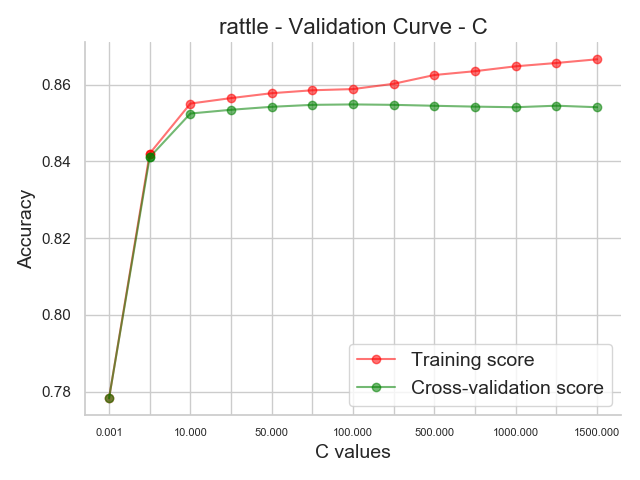}
\end{subfigure}
\hfill
\begin{subfigure}[b]{0.45\textwidth}
\centering
\includegraphics[width=\textwidth]{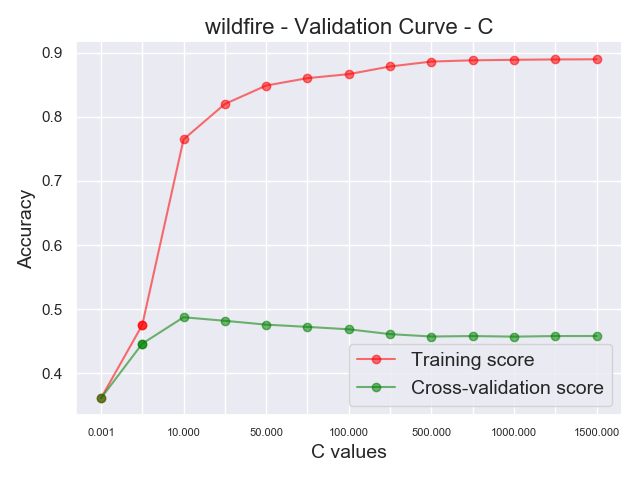}
\end{subfigure}
\caption{Effects of \textit{C} parameter on training/validation accuracy\label{fig:figure18}}
\end{figure}

\textbf{The \textit{C} parameter }represents the penalty of the error term and works as a regularizing component where with larger values, a tight decision boundary is preferred for better classification (overfit) while lower values support simple decision boundaries (underfit). This is evident from the plot where larger values indicate overfitting with the validation accuracy reducing with increase in C values. The low validation accuracy affirms the \textbf{high bias} with overfitting and the divergence confirms a\textbf{ large variance}. 

\textbf{The \textit{gamma} parameter }represents the kernel co-efficient and determines the circle of evaluation. With small values, a larger range of samples are used and with higher values, only a small set of closer samples are used, leading to overfitting.  This is evident from the plots where for lower values of gamma, the accuracy suffers with \textbf{low bias} while larger values tend to overfit leading to the curves diverging and displaying \textbf{high variance}. It is imperative to choose the appropriate C and gamma parameters to ensure the SVM algorithm performs optimally with lower bias and variance.

\begin{figure}[!htbp]
\centering
\begin{subfigure}[b]{0.45\textwidth}
\centering
\includegraphics[width=\textwidth]{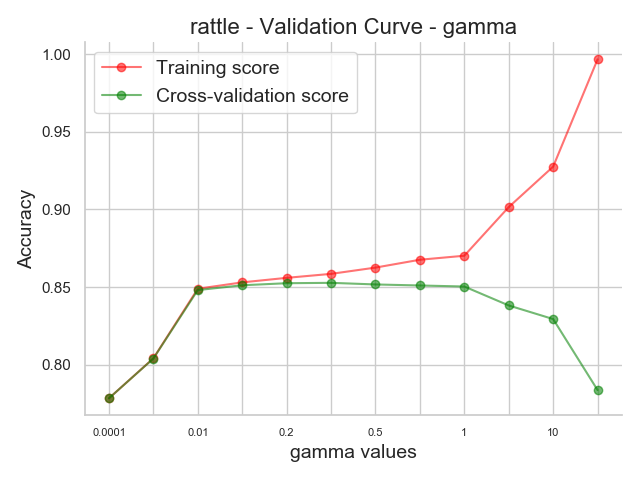}
\end{subfigure}
\hfill
\begin{subfigure}[b]{0.45\textwidth}
\centering
\includegraphics[width=\textwidth]{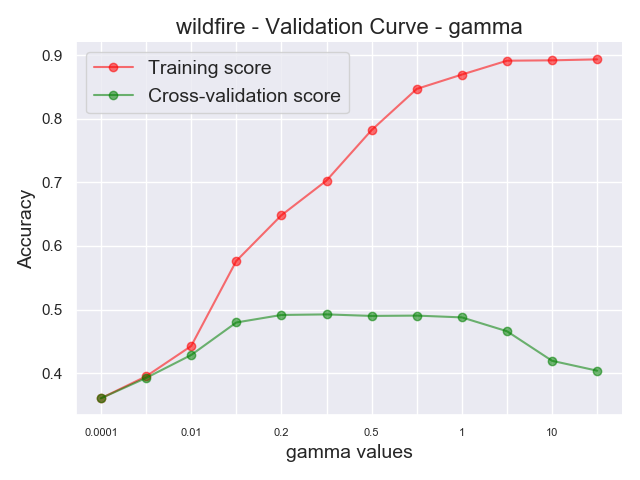}
\end{subfigure}
\caption{Effects of \textit{gamma} parameter on training/validation accuracy\label{fig:figure19}}
\end{figure}

Figure~\ref{fig:figure18} and Figure~\ref{fig:figure19} illustrate the effects of the two hyper parameters, C and \textit{gamma} on training and validation accuracies.

\subsubsection{Model Complexity Analysis with Kernel Comparison}

Using the optimal values the model complexity is analyzed over varying sizes of training samples using different kernels which determines the type of hyperplane used for separation of data. 

\begin{figure}[!htbp]
\centering
\begin{subfigure}[b]{0.45\textwidth}
\centering
\includegraphics[width=\textwidth]{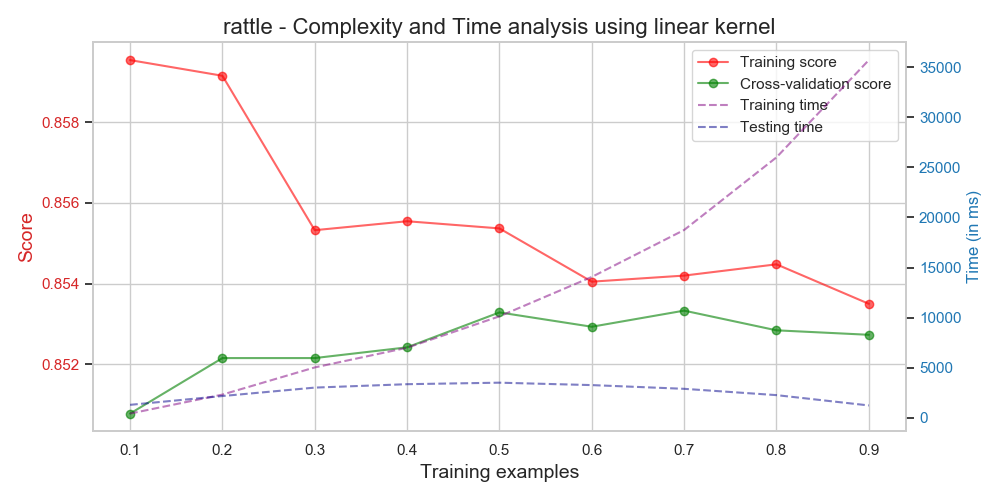}
\end{subfigure}
\hfill
\begin{subfigure}[b]{0.45\textwidth}
\centering
\includegraphics[width=\textwidth]{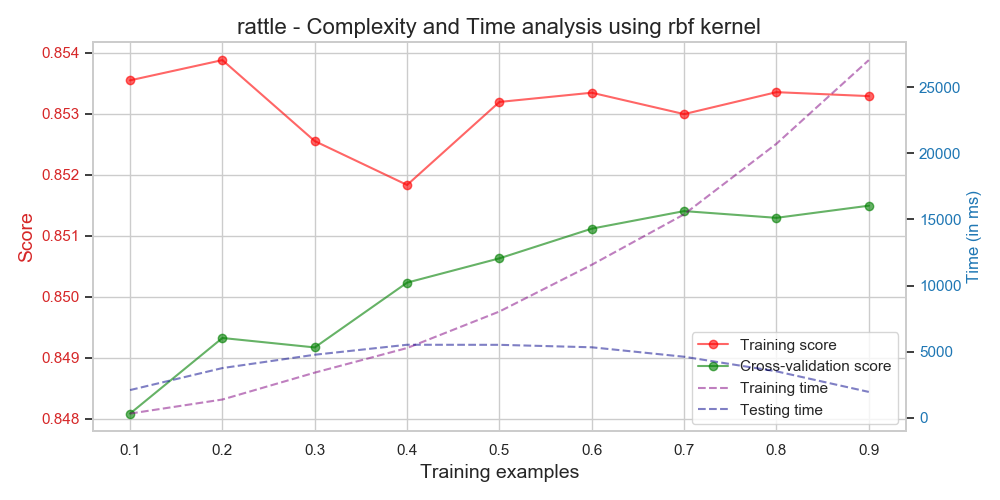}
\end{subfigure}
\caption{Learning curve analysis using \textbf{\textit{linear}} and \textbf{\textit{rbf}} kernels on Rattle dataset\label{fig:fugre20}}
\end{figure}

\begin{figure}[!htbp]
\centering
\begin{subfigure}[b]{0.45\textwidth}
\centering
\includegraphics[width=\textwidth]{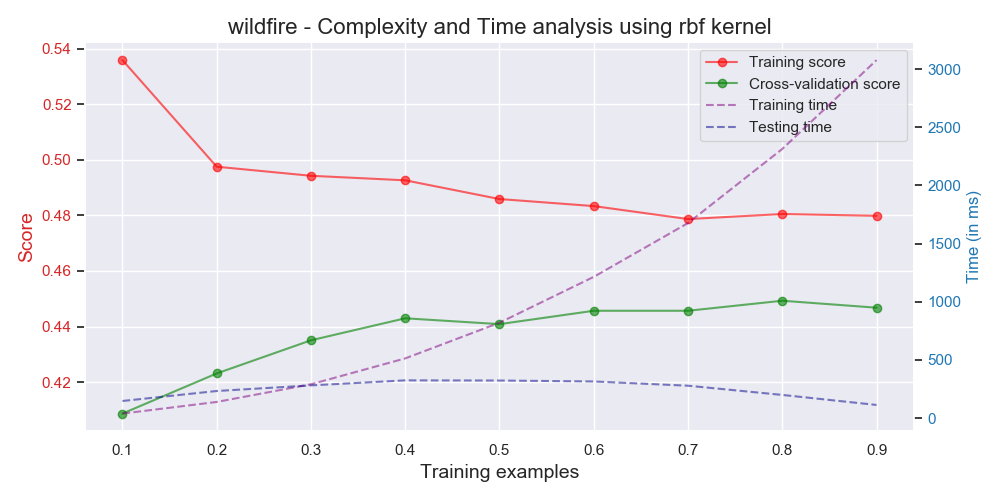}
\end{subfigure}
\hfill
\begin{subfigure}[b]{0.45\textwidth}
\centering
\includegraphics[width=\textwidth]{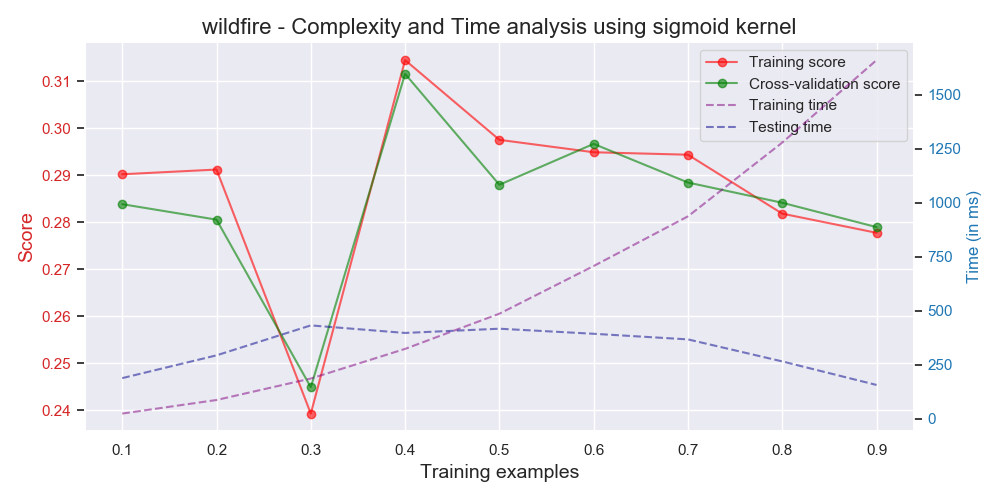}
\end{subfigure}
\caption{Learning curve analysis using \textbf{\textit{rbf}} and \textbf{\textit{sigmoid}} kernels on Wildfire dataset\label{fig:figure21}}
\end{figure}

Due to the richness of data and feature set in Rattle, the data was linearly separable allowing the linear kernel to function with slightly higher accuracy than the RBF kernel. As \cite{keerthi2003asymptotic} states, linear kernel is a degenerated version of RBF, and always has lesser accuracy than a tuned RBF kernel lending the argument that the RBF kernel may not have the most appropriate C and gamma parameters. However, due to its ability to work with non-linear boundaries, SVM exhibits a high range of accuracy and as evident from Figure~\ref{fig:figure21} has \textbf{low bias and very low variance} for both kernels. The training and validation scores do not vary on a large scale with training sample size and the curves almost converge displaying minimal variance. Due to its complexity, SVMs have larger training time with increase in training sample size as evident from the plots.

An interesting observation for the Wildfire dataset is the non-linear separability of its data, which might be due to the lack of features or insufficient data due to which the linear kernel does not converge even for 1000k iterations. However, RBF kernel capably sketches the nonlinear data in a higher dimensional space allowing SVM to separate the classes for better classification.

\begin{figure}[!htbp]
\centering
\begin{subfigure}[b]{0.45\textwidth}
\centering
\includegraphics[width=\textwidth]{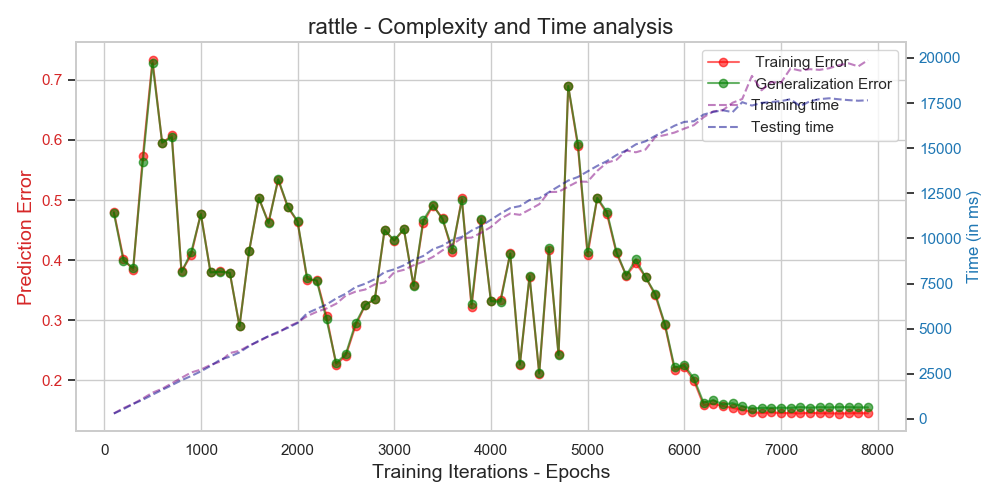}
\end{subfigure}
\hfill
\begin{subfigure}[b]{0.45\textwidth}
\centering
\includegraphics[width=\textwidth]{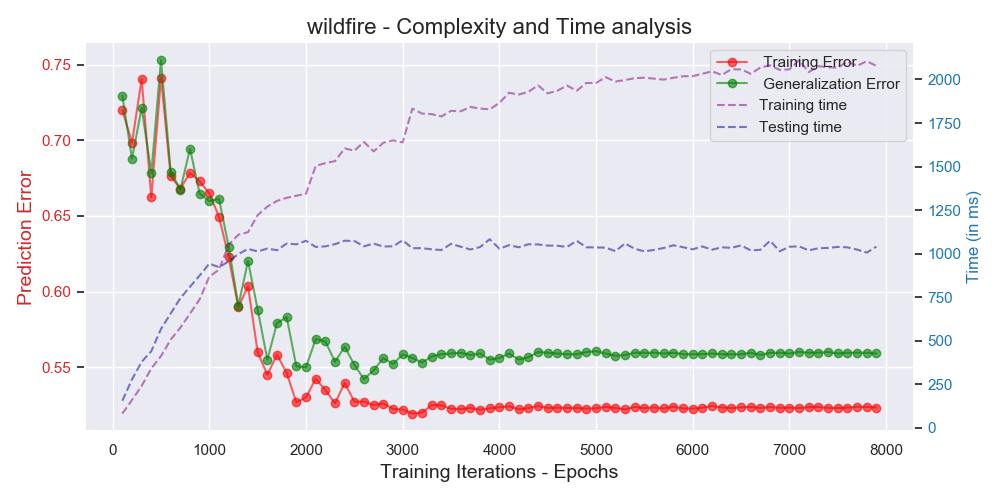}
\end{subfigure}
\caption{Learning curves for SVM for varied number of iterations/epochs\label{fig:figure22}}
\textit{Note: This learning curve is plotted with error metric in y-axis while others are plotted with accuracy}
\end{figure}

The iteration time (epochs) vs. error plot in Figure~\ref{fig:figure22} presents an interesting observation where \textbf{SVM exhibits good resistance to overfitting }even on large numbers of iterations, by maintaining the error rates at high epochs. The training and prediction time are linear to training epochs while after reaching a saturation remains constant as plotted for Wildfire. Both plots exhibit \textbf{low bias and very low variance}, presenting itself as the best model for these datasets.

\section{CONCLUSION}

We presented a simple evaluation of the classic algorithms on sparse tabular data for classification tasks and observed the effect on hyperparameters when the data is synthetically modified for higher noise. We were able to observe the efficiency of these algorithms in generalizing for sparse data and their utility of different parameters to improve classification accuracy. We were able to demonstrate that these classic algorithms are fair learners even for such limited data due to their inherent properties even for noisy and sparse datasets as observed in Table\ref{table:table4}.

Using the optimally tuned models, the accuracy was measured using the hold out test dataset and the results are available below in comparison with the default accuracy and their newly improved tuned accuracy.

\begin{table}[!htbp]
\renewcommand{\arraystretch}{1.3}
\begin{adjustbox}{max width=\textwidth}
\begin{tabular}{p{1.78cm}p{1.78cm}p{1.78cm}p{1.78cm}p{1.78cm}p{1.78cm}p{1.78cm}p{1.78cm}p{1.78cm}p{1.78cm}p{1.78cm}p{1.78cm}p{1.78cm}p{1.78cm}p{1.78cm}p{1.78cm}p{1.78cm}p{1.78cm}p{1.78cm}p{1.78cm}p{1.78cm}p{1.78cm}}
\hline
\multicolumn{1}{|p{1.78cm}}{\textbf{Algorithm}} & 
\multicolumn{2}{|p{3.56cm}}{\textbf{Decision Trees}} & 
\multicolumn{2}{|p{3.56cm}}{\textbf{Boosting}} & 
\multicolumn{2}{|p{3.56cm}}{\textbf{\textit{k-}Nearest Neighbors}} & 
\multicolumn{2}{|p{3.56cm}}{\textbf{Neural Network}} & 
\multicolumn{2}{|p{3.56cm}|}{\textbf{SVM}} \\ 
\hline
\multicolumn{1}{|p{1.78cm}}{\textbf{Rattle}} & 
\multicolumn{1}{|p{1.78cm}}{0.79} & 
\multicolumn{1}{|p{1.78cm}}{0.84} & 
\multicolumn{1}{|p{1.78cm}}{0.80} & 
\multicolumn{1}{|p{1.78cm}}{0.85} & 
\multicolumn{1}{|p{1.78cm}}{0.83} & 
\multicolumn{1}{|p{1.78cm}}{0.83} & 
\multicolumn{1}{|p{1.78cm}}{0.85} & 
\multicolumn{1}{|p{1.78cm}}{0.86} & 
\multicolumn{1}{|p{1.78cm}}{0.85} & 
\multicolumn{1}{|p{1.78cm}|}{0.85} \\ 
\hline
\multicolumn{1}{|p{1.78cm}}{\textbf{Wildfire}} & 
\multicolumn{1}{|p{1.78cm}}{0.47} & 
\multicolumn{1}{|p{1.78cm}}{0.55} & 
\multicolumn{1}{|p{1.78cm}}{0.55} & 
\multicolumn{1}{|p{1.78cm}}{0.59} & 
\multicolumn{1}{|p{1.78cm}}{0.51} & 
\multicolumn{1}{|p{1.78cm}}{0.49} & 
\multicolumn{1}{|p{1.78cm}}{0.44} & 
\multicolumn{1}{|p{1.78cm}}{0.51} & 
\multicolumn{1}{|p{1.78cm}}{0.50} & 
\multicolumn{1}{|p{1.78cm}|}{0.46} \\ 
\hline
\end{tabular}
\end{adjustbox}
\caption{Accuracy results of the algorithms using the default and tuned parameter values\label{table:table4}}
\end{table}

For the Rattle dataset, due to its large feature set and sample size, almost all algorithms functioned well with some hyper parameter tuning while for the Wildfire dataset with its mostly categorical features and sparse dataset, kNN and Decision Trees worked out really well with the highest accuracy values. It aligns with the understanding that large features and higher dimensionality fares well with ANN and SVMs while lower dimensional datasets perform better with Decision trees and more with Boosting. Though the Rattle dataset has a large feature set and training data, the accuracy still suffers from the curse of dimensionality and PCA or feature reduction would be required to add more weightage to high variance features. 

A random classifier has 0.66 and 0.27 accuracies for Rattle and Wildfire dataset exhibiting them as good datasets to experiment upon using these algorithms. All the algorithms performed better than the random classifier though the Wildfire dataset was more resistant to tuning due to its lack of features and missing data. There are indications of better performance with decision trees and using Boosting with almost 0.59 accuracy. Further analysis with more dimensionality and data can definitely improve the models’ performance.

\subsection{Further improvements and experiments}

Further analyses by changing the distance metric for KNN and using dimensionality reduction such as PCA to help optimize the distance metric, can be performed. Rebalancing some of the classes in the Wildfire dataset using undersampling, oversampling  or weight balancing to analyze their impact on accuracy can be attempted. Utilizing a few of the advanced neural networks, SVMs and deep learning models to experiment with their performance can yield more interesting results.

\printbibliography
\nocite{*}
\end{document}